\documentclass[lettersize,journal]{IEEEtran}

\usepackage[colorlinks,urlcolor=blue,linkcolor=blue,citecolor=blue]{hyperref}

\usepackage{color,array}

\usepackage{graphicx}
\usepackage{amsmath}
\usepackage{amssymb}
\usepackage{booktabs}
\usepackage{soul}
\soulregister\cite7
\soulregister\ref7
\soulregister\item7
\soulregister\itemize7
\soulregister\pageref7
\usepackage{multirow}
\usepackage{algorithm}
\usepackage{algorithmic}
\usepackage{caption}
\usepackage{subcaption}

\makeatletter %% <- make @ usable in command sequences
\newcount\SOUL@minus
\makeatother  %% <- revert @

\begin{document}

\title{Towards Efficient Convolutional Neural Networks with Structured Ternary Patterns}

\author{Christos Kyrkou
        % <-this % stops a space
\thanks{Author Accepted Manuscript. Final version appears in Transactions on Neural Networks and Learning Systems. (\url{https://doi.org/10.1109/TNNLS.2024.3380827})}
\thanks{The author is with the KIOS Research and Innovation Center of Excellence, University of Cyprus, Nicosia, Cyprus (email: kyrkou.christos@ucy.ac.cy))}% <-this % stops a space
}

\markboth{}%
{Shell \MakeLowercase{\textit{et al.}}: A Sample Article Using IEEEtran.cls for IEEE Journals}

\IEEEpubid{}

\maketitle

\begin{abstract}
High efficiency deep learning models are necessary to facilitate their use in devices with limited resources but also to improve resources required for training. Convolutional neural networks typically exert severe demands on local device resources and this conventionally limits their adoption within mobile and embedded platforms. This paper presents work towards utilizing static convolutional filters generated from the space of local binary patterns and Haar features to design efficient convolutional neural network architectures. These are referred to as \textbf{S}tructured \textbf{T}ernary \textbf{P}atterns (STeP) and can be generated during network initialization in a systematic way instead of having learnable weight parameters thus reducing the total weight updates. The ternary values require significantly less storage and with the appropriate low level implementation, can also lead to inference improvements. The proposed approach is validated using four image classification datasets, demonstrating that common network backbones can be made more efficient and provide competitive results. It is also demonstrated that it is possible to generate completely custom STeP-based networks that provide good trade-offs for on-device applications such as Unmanned Aerial Vehicle based aerial vehicle detection. The experimental results show that the proposed method maintains high detection accuracy while reducing the trainable parameters by 40-80\%. This work motivates further research towards good priors for non-learnable weights that can make deep learning architectures more efficient without having to alter the network during or after training.
\end{abstract}


\section{Introduction}
\IEEEPARstart{D}{eep} learning (DL) and artificial intelligence (AI) are rapidly being deployed in on-device scenarios, such as autonomous driving, robotic systems, and unmanned aerial vehicles \cite{Kyrkou2021}. This proliferation can be attributed to several factors, including privacy concerns, the need for real-time decision-making, and the necessity to address connectivity limitations. Despite the adoption of deep learning models in mobile applications, achieving consistently high-performance inference on mobile and embedded platforms remains a formidable challenge, primarily due to the substantial computational requirements involved. In pursuit of enabling on-device AI capabilities, researchers are constantly seeking strategies aimed at enhancing the computational efficiency of inference processes \cite{Zhang_2018_CVPR_Workshops}.

Thus, the community has strived to identify fundamental components suitable for utilization in deep learning algorithms, specifically convolutional neural networks (ConvNets). Techniques such as network pruning and parameter compression/quantization are increasingly gaining prominence as strategies to facilitate the deployment of deep learning models within fixed-point pipelines, as shown in \cite{MobileAIW}. However, not all existing network architectures readily accommodate quantization, as this transformation may introduce a significant accuracy gap when compared to their floating-point counterparts. Furthermore, it can be challenging to anticipate the a priori impact of optimization techniques on both accuracy and the ensuing computational complexity. Consequently, in many instances, the training process becomes more intricate, necessitating iterative refinement.

\begin{figure}[t]
\centering
\includegraphics[width=0.99\columnwidth]{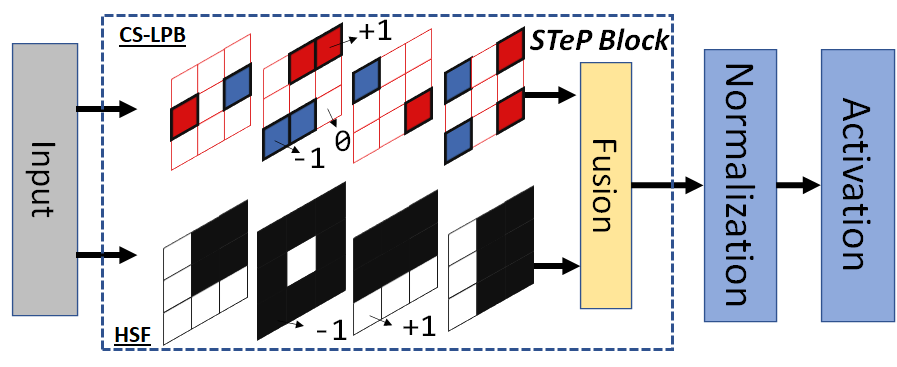}
\caption{Basic STeP Block: Filters are composed of center symmetric local binary patterns (CS-LBP) and Haar-structured features (HSF). Both use ternary values necessitating only addition/subtraction operations.}
\label{fig:basic_block}
\end{figure}

This work explores the utilization of \textbf{S}tructured \textbf{Te}rnary \textbf{P}atterns (STeP) as ConvNet weights,  approximating Haar features and center-symmetric local binary patterns, as a viable alternative to full-precision trainable weights. Thus establishing a competitive baseline for efficient convolutional neural networks utilizing ternary values ($[-1,0,1]$). The primary objective is to enable direct training with these structured weights, leading to the creation of streamlined models, which hold the potential for efficiency improvements in both inference and training processes. The present work introduces a method for the pre-generation of such structured ternary patterns, which can be seamlessly integrated into existing neural networks (e.g., ResNet) or employed to construct efficient custom ConvNets for on-device applications. The key advantages of this approach are summarized as follows:

\noindent• Does not necessitate any change to existing training procedures or post-retraining.\\
• Reduces the number of learnable parameters, thereby enabling the utilization of larger batches due to the reduced storage requirements during the backward propagation phase.\\
• Mitigates the storage demands for weights by utilizing lower-precision ternary values.\\
• Offers the potential for multiplier-free operations in the case of ternary weight kernels.

Results for classification and object detection tasks, across various datasets demonstrate the effectiveness of the approach. The proposed method achieves a substantial reduction in the number of trainable parameters, ranging from $40-80\%$ while experiencing only a marginal average accuracy reduction of $3.1\%$, achieved by utilizing ternary weights instead of floating-point parameters. When combined with a compact model architecture, it yields models that are $40\times$ times smaller in size while maintaining competitive performance. These parameter savings are accompanied by comparable accuracy results and do not necessitate post-training network modifications.

\section{Related Work}

\subsection{Binarized Neural Networks}
Works such as \cite{courbariaux2016binarized} demonstrate a methodology that utilizes full-precision latent weights during the training process to accumulate gradients. Subsequently, these accumulated gradients are employed to compute the corresponding binary weights. Building upon this foundation, other studies have extended the exploration of various properties, including scaling factors, for binarized networks \cite{XORNet}. Binary weight neural networks exhibit an advantageous compression ratio; however, they often incur a substantial sacrifice in accuracy when compared to baseline full-precision networks. Techniques like those introduced in \cite{bitE} and \cite{Abdolrashidi_2021_CVPR} necessitate additional training for weight quantization, typically restricting weights to fixed-point representations. Alternatively, rather than training a full-precision network followed by quantization/binarization, alternative research avenues delve into the design of binary architectures. For instance, the work in \cite{ijcai2018-380} dissects convolutional kernels into binary basis functions, generating filters as a linear combination of orthogonal binary codes in real-time. This transformation in computation, however, does not alleviate the count of learnable parameters. 
In a similar fashion, \cite{LBPconv} introduces local binary convolutional networks that replace traditional convolutional layers with a learnable module inspired by local binary patterns. This approach necessitates learning pivot point locations and incorporates a non-linear thresholding operation, which can influence the flow of gradients. Similarly, \cite{LBPN} implements the local binary operator as a sampling operation. This involves learning the points to generate multiple output channels, each equipped with its own binary output code. Finally, \cite{TNNLS5} demonstrates the benefits of software-hardware codesign for extremely sparse networks with binary connections for image classification. Such accelerators can also be utilized for the proposed work as well. Despite the promising advantages that binary networks offer over their full-precision counterparts, vanilla binary networks frequently suffer from substantial accuracy degradation, as highlighted in \cite{courbariaux2016binarized}. Furthermore, existing schemes fail to reduce the overall number of learnable parameters and necessitate gradient tracking for all parameters, as pointed out in \cite{zhu2016trained}.

\subsection{Quantization of Neural Networks}
Quantization, in general, is a method employed to map input values from a large, often continuous set to output values within a smaller, often finite set. Examples of quantization techniques include rounding and truncation \cite{MobileAIW}. Within the context of neural networks, quantization plays a pivotal role in enabling the conversion of network weights and activation functions from floating-point operations to fixed-point or mixed-precision operations. This transition not only reduces the model's storage footprint and memory consumption but also enhances both latency and power efficiency \cite{liang2021pruning}. Traditionally, uniform quantization, where all layers of the network share the same bit-width, has been utilized \cite{jacob2018quantization}. More recently, adaptive deployment strategies, which involve varying the bitwidth of weights and intermediate activations, have emerged as an alternative approach \cite{jin2020adabits}. \cite{TNNLS1_bit_assign} demonstrates optimal bitwidth assignment for weight and activation quantization of deep convolutional neural networks using dynamic programming. Similarly, \cite{TNNLS6} focuses on image super-resolution applications using dynamic thresholds for quantization. Ternary quantized networks have been demonstrated for federated learning settings in \cite{TNNLS3_tcompress_FL}. It is worth noting that while quantization techniques offer substantial benefits, they are often accompanied by a noticeable decline in model accuracy, necessitating re-training in many cases \cite{ghamari2021quantization}. Nevertheless, these techniques can be applied independently of the weight and architectural modifications proposed in this paper, potentially yielding further improvements. This synergy is elaborated further in Section \ref{sec:ODexp}.

\subsection{Efficient Neural Networks}
One critical aspect of designing efficient neural networks pertains to their architectural considerations. A multitude of works have tackled this challenge, with a primary focus on architectural and layer modifications aimed at optimizing the size and speed of deep learning models \cite{menghani2023efficient}. Notably, experts in the field have sought to engineer lightweight, parameter-efficient networks, introducing significant changes at both the micro- and macro- architecture levels \cite{ZewenSurveyCNN2022}. These architectural adjustments encompass various techniques, such as the decomposition of convolutional layers, as exemplified by Wang et al. \cite{wang2018pelee}, who advocated splitting a $3\times3$ kernel into two consecutive layers employing $1\times3$ and $3\times1$ kernels, respectively. Another innovation, demonstrated by Koonce et al. \cite{koonce2021squeezenet}, involves the utilization of "fire modules" that employ 1x1 convolutions to compress parameters efficiently. Additionally, the introduction of "bottleneck" blocks has gained attention, as seen in MobileNetV2 \cite{MobileNetV2}, which employs two convolutional layers with a $1\times1$ kernel to reduce the dimensionality of features processed by larger filters (e.g., $3\times3$). This concept is further refined in Ma et al.'s work \cite{ma2018shufflenet}, where convolutional layers are decomposed into a sequence of smaller, independent layers using grouped convolutions. EfficientNet, introduced by Tan et al. \cite{EfficientNet:2019}, stands as a notable convolutional neural network architecture and scaling method. It adopts a compound coefficient to uniformly scale the depth, width, and resolution dimensions of the network, thereby achieving efficiency across various dimensions. Another noteworthy innovation is the "ghost modules," as introduced by Han et al. \cite{han2020ghostnet}. These modules are designed to generate more features while using fewer parameters, thereby enhancing network efficiency. It is important to note that architectural modifications can be seamlessly integrated with other techniques to further enhance the overall efficiency of neural networks.

\section{Proposed Approach}

\begin{figure}[t]
\centering
\includegraphics[width=0.9\linewidth]{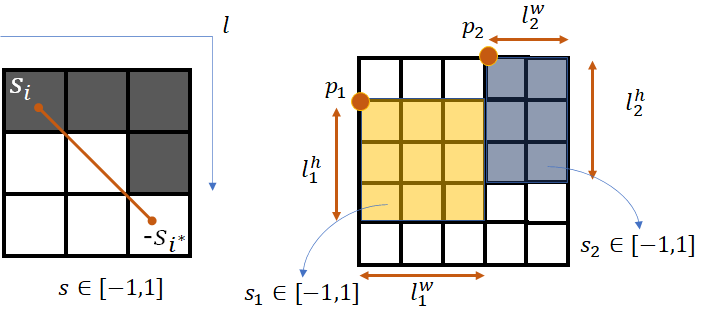}
\caption{(left) Parametrization of the CS-LBP features. (right) Parametrization of Rectangular Haar Features with two boxes.}
\label{fig:feature_params}
\end{figure}

\subsection{Generating Static Structured Patterns}
The proposed approach involves the generation of convolutional filter weights based on structured patterns that are known to exhibit exceptional performance in computer vision tasks, such as in the domains of face and pedestrian detection \cite{SHETTY2021330,6909521}. The utilization of predetermined binary/ternary filters offers several notable advantages, primarily the potential reduction of memory and computational requirements during both training and inference phases. Specifically, this approach leverages two well-established pattern extractors: Local Binary Patterns (LBPs) \cite{CSLBP:2010,Kyrkou:TVLSI:2016} and Haar-like features \cite{990517,Kyrkou:TVLSI:2010}. LBPs have proven their versatility across a broad spectrum of applications for constructing robust visual object detection systems. In this research, Center-Symmetric Local Binary Patterns (CS-LBP) \cite{CSLBP:2010,Kyrkou:JPP:2018} are utilized, which exhibit enhanced resilience to flat regions and a heightened ability to capture gradient information compared to the original LBP formulation (as employed in \cite{LBPconv}), all without necessitating the use of any thresholds. Moreover, the approach introduces the incorporation of Haar-based features, which had a seminal impact on real-time object detection \cite{990517}. This integration represents a novel exploration within this context. Herein, these two distinct feature families are synergistically employed as the foundation for generating convolutional kernels that possess specific structural patterns. Notably, these kernels only necessitate ternary representation values of $[-1, 0, 1]$, thus making them especially efficient in terms of computation and memory, requiring only addition/subtraction for the ternary weight kernels.

Other approaches, such as \cite{LBPCOnvNet2017:cvpr}, share a common foundation inspired by the original Local Binary Pattern (LBP) formulation. However, they diverge in their approach by randomly initializing the convolutional kernel values within the range of $[-1,0,1]$ while imposing only sparsity constraints and lacking structural considerations. In contrast to the stochastic initialization utilized in \cite{LBPCOnvNet2017:cvpr}, the proposed methodology explores a more structured parameter space. Specifically, weights from the spaces of Center-Symmetric Local Ternary Patterns (CS-LBP) and Haar-Structured Features (HSF) are sampled to attain a more principled and discriminative representation.

\begin{algorithm}[t]
	\caption{Generate CS-LBP Features}
	\label{alg:cslbp}
	\begin{algorithmic}[1]
	\STATE \textbf{Inputs: } number of channels: $k$, width: $w$, height: $h$, output filters: $f$\\
	\STATE \textbf{Output: } Initialized kernels
	
		\STATE \textbf{function} \textit{Generate\_CS\_LBP}($weights$):
		\STATE $weights$ $\leftarrow$ Tensor($h,w,k,f$) 
		\FOR{\textit{every} weight\_matrix in \textit{layer}} 
		\STATE length$\leftarrow$ RandomInt(0,4)
		\FOR{\textit{i} $\rightarrow$ [0,...length]}
		\STATE $s$ $\leftarrow$ RandomFromSet($[-1,0,1]$)
		\STATE $i* \leftarrow i$ \textbf{\%Opposite value index}
		\STATE weight\_matrix[i] $\leftarrow$  $s$
		\STATE weight\_matrix[i*] $\leftarrow$  $-s$
		\ENDFOR
		\STATE UPDATE $weights$ with $weight\_matrix$ 
		\ENDFOR
		\STATE \textbf{return} $weights$

	\end{algorithmic}
\end{algorithm}

\begin{algorithm}[t]
	\caption{Generate Rectangular Haar Features}
	\label{alg:haar}
	\begin{algorithmic}[1]
	\STATE \textbf{Inputs: } number of channels: $k$, width: $w$, height: $h$, output filters: $f$\\
	\STATE \textbf{Output: } Initialized kernels
	
		\STATE \textbf{function} \textit{Generate\_CS\_LBP}($weights$):
		\STATE $weights$ $\leftarrow$ Tensor($h,w,k,f$) 
		\FOR{\textit{every} weight\_matrix in \textit{layer}} 
		\STATE $r \leftarrow$ RandomInt(1,4) \textbf{\%Number of rectangles}
		\FOR{$i$ \: 2 to \textit{r}} 
		\STATE $p_i\leftarrow$ [RandomInt(0,$w$-1),RandomInt(0,$h$-1)]
		\STATE $l_i^w\leftarrow$ RandomInt(0,$w$-1)
		\STATE $l_i^h\leftarrow$ RandomInt(0,$h$-1)
		\STATE $s_i$ $\leftarrow$ RandomFromSet($[-1,1]$) 
		\STATE weight\_matrix[$p_i$:$l_i^w$,$p_i$:$l_i^h$] $\leftarrow$  $s_i$
		\ENDFOR
		\STATE UPDATE $weights$ with $weight\_matrix$ 
		\ENDFOR
		\STATE \textbf{return} $weights$
	\end{algorithmic}
\end{algorithm}

\begin{figure*}[t!]
    \centering
    \begin{subfigure}[t]{0.45\textwidth}
        \centering
        \includegraphics[width=\textwidth]{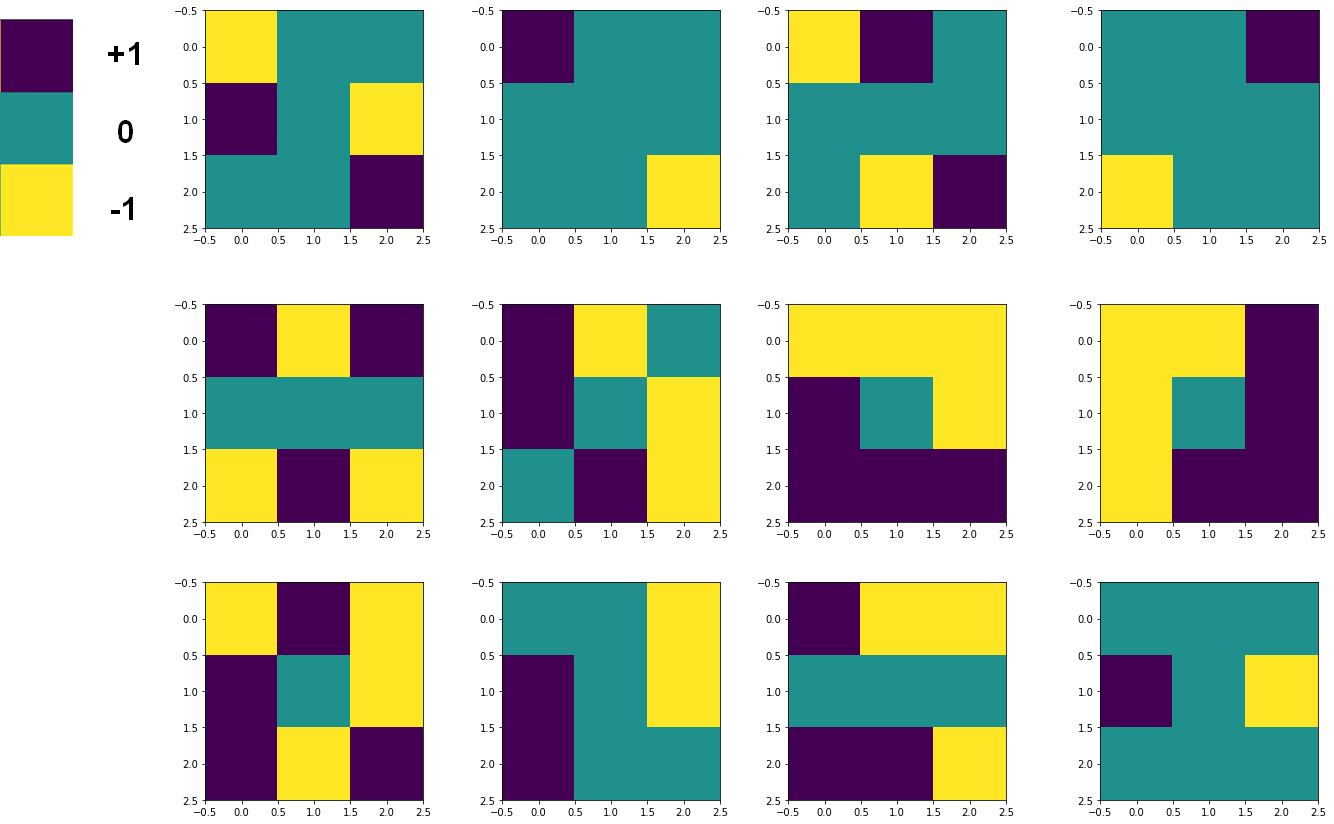}
     \caption{Center-symmetric local ternary weights of size $3\times3$.}
     \label{fig:cs_features}
    \end{subfigure}%
    ~ 
    \begin{subfigure}[t]{0.45\textwidth}
        \centering
        \includegraphics[width=0.85\textwidth]{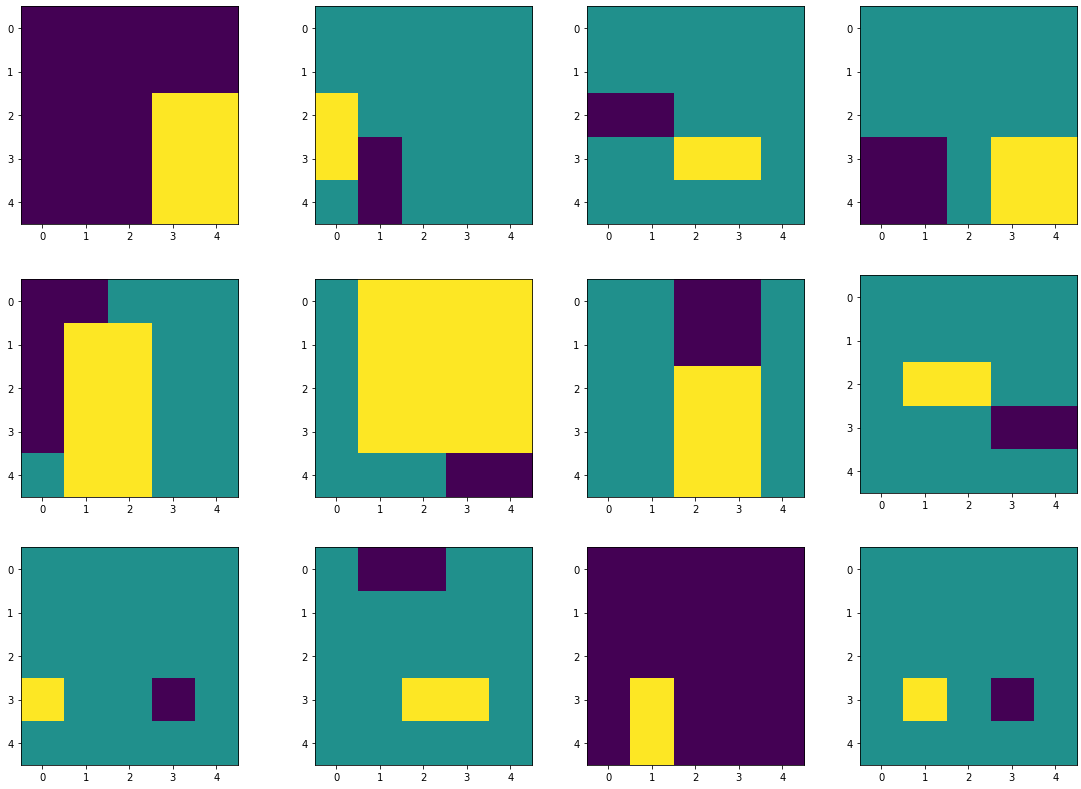}
     \caption{Haar-based ternary weights of size $5\times5$}
     \label{fig:haar_features}
    \end{subfigure}
   \caption{Illustration of static generated ternary features by specifying different parameters.}
    \label{fig:all_features}
\end{figure*}

CS-LBP patterns rely on the differences between diagonal center-symmetric pairs of kernel values, as illustrated in Fig. \ref{fig:feature_params}, and are closely related to the gradient operator. They are adept at detecting corners and edges while remaining unaffected by the central location of the patterns. Additionally, CS-LBP patterns can take on zero values, making them inherently more sparse. HSF filters, on the other hand, operate by considering differences between rectangular regions, as depicted in Fig. \ref{fig:feature_params}. Values belonging in the same region are accumulated together, enabling them to detect more abstract patterns and flat regions. The specific type of Haar feature, along with the random design of its location and weight for each rectangular region, allows for a high degree of adaptability in capturing diverse features and textures.

The generated patterns undergo convolution with the input tensor, utilizing a methodology akin to the conventional convolution operator. Notably, this approach offers the added advantage of restricting operations solely to addition and subtraction. The computation of each feature is shown in Eq. \ref{eq:lbp} and \ref{eq:hsf}. The computation of each feature happens as shown in Eq. \ref{eq:lbp} and \ref{eq:hsf}. In the context of CS-LBP $N$ denotes the count of pairs of opposing elements within the kernel, $s_i$ signifies the sign associated with each such pair, and $a_i$ is a binary indicator denoting whether a given pair contributes to the kernel or not. Conversely, for the case of HSF $M$ represents the total number of elements in the kernel and $s_i$ characterizes the sign assigned to each individual element in the kernel. 

\begin{gather}
CS-LBP = \sum_{i=1}^{N}{a_i(s_ix_1^i-s_ix_2^i)}
\label{eq:lbp}
\end{gather}

\begin{gather}
HSF = \sum_{i=1}^{M}{s_ix^i}
\label{eq:hsf}
\end{gather}

Features are generated prior to training in a channel-wise fashion through parameterization of distinct properties within each feature space, as illustrated in Fig. \ref{fig:feature_params}. The weights for CS-LBP are computed by iteratively sampling values from the interval $[-1, 1]$ for half of the weight locations and subsequently inverting the sign for the mirrored locations, as depicted in Fig. \ref{fig:feature_params}. Haar features are constructed based on the number of rectangles within the feature, typically ranging from 2 to 4 (Alg. \ref{alg:haar}, Line 4). The top-left position of these rectangles is randomly selected within the weight matrix, and their side lengths are sampled uniformly according to provided parameters (Alg. \ref{alg:haar}, Lines 6-8). It is important to note that these values are adjusted to ensure compatibility with the size of the weight matrix; however, for the sake of clarity, the specific adjustments are omitted. The detailed procedures for generating these features are outlined in Alg. \ref{alg:haar} and Alg. \ref{alg:cslbp}, respectively. In these algorithms, the kernels are initially initialized with zero values and are subsequently modified in accordance with the defined procedures. This approach yields a diverse set of features capable of processing both coarse and fine-grained patterns, as shown in Fig. \ref{fig:all_features}.

\subsection{Networks Composed of Structured Ternary Patterns}
\textbf{Basic Block:} The proposed approach is capable of generating weight patterns based on underlying data structures, which can be utilized to construct a fundamental building block referred to as \textbf{STeP} (\textbf{S}tructured \textbf{Te}rnary \textbf{P}attern). This block can be seamlessly integrated into both classification and detection networks. The formation of this block involves an initial step of calculating outputs from the convolution process using both CS-LBP weights and HAAR features. Subsequently, these outputs can be combined either through element-wise addition or via a $1\times1$ convolution operation, which itself can have weight values within the range of $[-1,1]$. Furthermore, the block incorporates pooling, normalization, and activation functions after the feature fusion process. This primary block can be directly integrated into standard ConvNet architectures to replace computationally expensive operations.

\noindent\textbf{Custom network:} The primary objective of the basic block is to facilitate the construction of custom networks that incorporate the STeP block. In the context of these custom networks,  batch normalization and LeakyReLU activations are employed within the STeP block. During the testing phase, it is possible to optimize the LeakyReLU activation function by zeroing out its negative components and rounding the positive values to the nearest integer. This refinement serves to enhance the model's hardware compatibility and efficiency. Consequently, the CS-LBP/HSF block can consistently be delegated to a hardware accelerator for streamlined processing, while also reducing the computational load on the CPU. Additionally, a detection layer can be incorporated into the architecture to perform tasks such as bounding box regression and object classification. It's important to note that non-binary operations, such as batch normalization with scaling and shifting, remain in floating-point precision. Similarly, the activation functions have not undergone any alterations.

\section{Experiments and Results}
The conducted experiments encompass both classification and detection tasks, with the primary objective of demonstrating the versatile utility of the STeP blocks. Specifically, this research seeks to establish that these STeP blocks can serve a dual purpose: 1) enhancing the computational efficiency of pre-existing network architectures by seamlessly integrating STeP blocks in lieu of traditional components, and 2) empowering the creation of custom neural networks tailored for real-world applications, thereby achieving a balance between competitive accuracy and efficiency improvements.

\subsection{Image Classification}

\begin{table*}[t]
\begin{center}
\caption{Comparison of Classification Networks\\}\label{tb:compArch}
\resizebox{\textwidth}{!}{%
    \begin{tabular}{c|c|c|c|c|c|c|c|c|c} 
      \textbf{Model} & \textbf{Type} & \textbf{Trainable} & \textbf{Non Trainable} & \textbf{Acc-C10} & \textbf{Acc-C100} & \textbf{Acc-ImgN16} & \textbf{Acc-Tiny200} & \textbf{Memory}$^\dag$ & \textbf{Reduction}$^\dag$ \\ 
      \textbf{Family} & & \textbf{Parameters}$^\dag$ & \textbf{Parameters}$^\dag$ & \textbf{(\%)} & \textbf{(\%)} & \textbf{(\%)} & \textbf{(\%)} & \textbf{(MB)} & \textbf{(\%)} \\ 
      \hline
      \textit{VGG-16} & STeP(Prop.) & $1,652,490$ & $13,934,754$ & $90.4$ & $66.6$ & $81.1$ & $54.0$ &  $10.1$  & $82.8$ \\
      \cite{VGG} & Original & $14,728,266$ & $-$ & $94.0$ & $74.2$ & $85.3$ & $62.2$ &  $58.9$ & $-$ \\
      & Rand. Binary & $13,578$ & $14,710,464$ & $61.8$ & $35.6$ & $61.3$  & $24.1$ & $1.89$  & $96.7$\\
      \hline
      \textit{ResNet50} & STeP(Prop.) & $13,463,114$ & $11,318,976$ & $94.8$ & $77.9$ & $91.2$  & $68.1$ &  $56.6$  & $39.8$ \\
      \cite{ResNet}& Original & $23,520,842$ & $-$ & $95.4$ & $78.0$ & $88.2$ & $69.2$ &  $94.1$  & $-$ \\
      & Rand. Binary & $12,201,866$ & $11,318,976$ & $90.7$ & $67.6$ & $78.5$  & $57.9$ & $51.6$  & $45.1$ \\
      \hline
      \textit{MobileNet} & STeP(Prop.) & $1,268,858$ & $1,091,810$ & $91.4$ & $69.1$ & $89.6$  & 57.0 & $5.3$  & $37.3$\\
      V2 & Original & $2.360.668$ & $-$ & $94.3$ & $72.8$ & $91.0$ & 60.6 & $9.4$ & -  \\
      \cite{MobileNetV2} & Rand. Binary & $859,034$ & $1,437,888$ & $69.1$ & $59.4$ & $70.1$  & 48.9 & $3.6$  & $61.7$ \\
      \hline
      \textit{EfficientNet} & STeP(Prop.) & $1,932,806$ & $1,644,578$ & $88.9$ & $63.4$ & $83.5$ & 53.0 & $8.1$  & $45.3$ \\
      B0 & Original & $3,598,598$ & $-$ & $91.4$ & $69.8$ & $85.6$ & 58.7 &  $14.3$ & -  \\
      \cite{EfficientNet:2019} & Rand. Binary & $1,932,806$ & $1,563,938$ & $82.0$ & $55.7$ & $33.5$ & 51.1 & $7.9$  & $44.7$ \\
      \hline
    \end{tabular}
	}
     \label{tab:models_acc_cls}
\end{center}
    $^\dag$ Numbers calculated based on the architecture generated for the CIFAR-10 dataset. They are very similar for other datasets.
\end{table*} 

To assess the effectiveness of the proposed approach, experiments are conducted using four distinct classification datasets. The CIFAR-10 dataset (referred to as C-10) \cite{krizhevsky2009learning} consists of $32\times32$ RGB images, containing 50,000 images for training and 10,000 images for testing, distributed across ten classes. The CIFAR-100 dataset (referred to as C-100) encompasses a more complex classification task, comprising 100 distinct classes. This dataset consists of 600 images per class, divided into 500 training images and 100 testing images, all with a resolution of $32\times32$ pixels. In addition, the Tiny-ImageNet dataset (referred to as Tiny200), it also utilized, which serves as a subset of the larger ImageNet dataset \cite{deng2009imagenet}. Tiny200 contains 100,000 images distributed across 200 classes, with each class containing 500 images. These images have been resized to $64\times64$ pixels, offering a balance between image quality and computational complexity. Lastly, the Imagenet-16\footnote{\url{https://zenodo.org/record/8027520}} dataset (referred to as ImgN16) is another subset of the renowned ImageNet dataset. It features a more compact classification task with 16 distinct classes. Notably, ImgN16 maintains the original image resolutions, which are typically resized to $224\times224$ pixels for standard ImageNet evaluations. Collectively, these datasets provide a comprehensive evaluation set, encompassing a wide range of class counts and image resolutions, thereby enabling a thorough assessment of the proposed approach.

Four prominent network architectures in the field of computer vision are used in the evaluation with distinct characteristics and performance. The models considered are the ResNet50 \cite{ResNet}, MobileNetV2 \cite{MobileNetV2}, EfficientNet \cite{EfficientNet:2019}, and VGG16 \cite{VGG} which provide a comprehensive assessment across different design philosophies and trade-offs. The comparative analysis also encompasses three distinct weight generation settings. The original model utilizing weight initialization and update through backpropagation, models initialized with random binary weights, and models utilizing the STeP block with structured ternary weights.

It is worth noting that not all network blocks were converted to non-learnable ternary values. The changes concerned mainly the spatial kernels. Bottleneck layers applied to shortcut connections, squeeze and excitation layers, and the final classification layer weights were not converted for the purposes of these experiments. This is left as additional area of investigation, while benefits are still tangible even with partial modification.

All networks are trained from scratch for $200$ epochs, with a starting learning rate of $0.1$ and weight decay of $5\times10^{-4}$, and a cosine annealing scheduler. An initial batch size of 128 is used. Where feasible for the binary/STeP models larger batch sizes are applied. During the training process, the data are augmented with random padding and crop to standard size, as well as random horizontal flipping as per standard practice.

\begin{figure*}[t]
\centering
\includegraphics[width=0.8\linewidth]{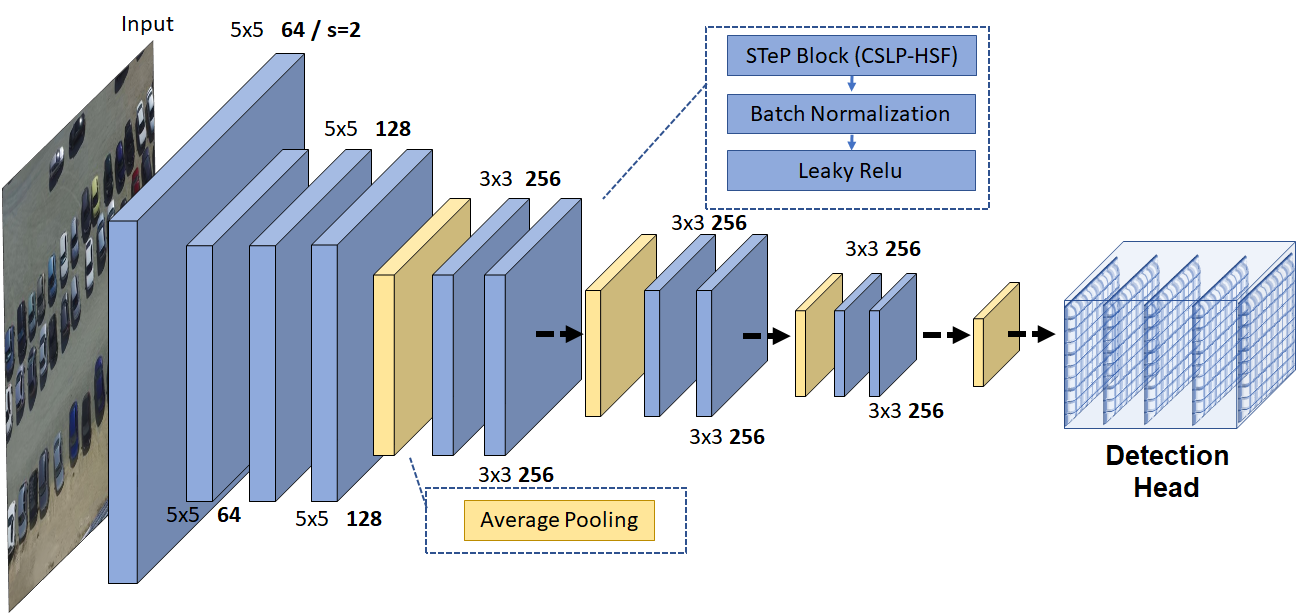}
\caption{STeP network architecture used for object detection.}
\label{fig:CNN_model}
\end{figure*}

The results, as presented in Table \ref{tab:models_acc_cls}, showcase the notable enhancements in model characteristics arising from the proposed methodology. It is noteworthy that the VGG-16 network exhibits the most substantial reduction in learnable parameters, primarily owing to its utilization of traditional convolution layers. Consequently, employing ternary values for storage leads to a $82\%$ reduction in storage size. As anticipated, the random binary weight model demonstrates the most significant reduction in learnable parameters; however, this improvement is accompanied by consistently diminished accuracy across all datasets. On average, the proposed approach demonstrates a commendable average reduction of approximately $56.1\%$ in learnable parameters and a noteworthy average memory reduction of approximately $50.8\%$ across various models.

With regard to attainable accuracy across all datasets, it is noteworthy that while random binary patterns exhibit competitive performance in certain instances, the STeP block consistently demonstrates superior performance across all experimental scenarios. More precisely, the STeP models exhibit only an average accuracy degradation of $3.1\%$ compared to the full-precision model, whereas the random binary models experience an average accuracy reduction of $20.2\%$. Notably, when examining specific cases, we find that the STeP model derived from ResNet even manages to yield a slight improvement in accuracy for the Imagenet-16 dataset. This phenomenon can be attributed to the reduced number of learned parameters, which mitigates overfitting to some extent. Conversely, in the same case, the random binary model experiences a significant decline of over $9\%$ compared to the original model. Across various model architectures, it is evident that STeP variants consistently deliver robust performance. Notably, the ResNet network experiences the least reduction in performance, primarily due to its larger number of trainable parameters. Even for datasets with larger image dimensions, such as Imagenet-16 and TinyImagenet, the STeP-based performance remains consistent. Also, compared to more complex approaches for creating efficient neural networks \cite{TNNLS4}, the proposed approach demonstrates improved performance on C-10 and C-100, and is extensible to object detection as well while it does not require iterative training.

\begin{table*}[t]
\begin{center}
\caption{Comparison of Detection Networks\\}\label{tb:compArch}
\resizebox{\textwidth}{!}{%
    \begin{tabular}{l|c|c|c|c|c} 
      \textbf{Backbone} & \textbf{Trainable} & \textbf{Non Trainable Param.} & \textbf{mAP} & \textbf{IoU} & \textbf{Memory(MB)} \\ 
      \hline
      \textit{STeP-Det (Proposed)} & $232,492$ & $3,738,560$ & $0.68$ & $0.63$ & $1.6$  \\
      \hline
      \textit{LBC-Net} \cite{LBPconv} & $455,326$ & $3,507,648$ & $0.61$ & $0.48$ & $2.2$  \\
      \hline
      MobileNet \cite{howard2017mobilenets}
 & $3,477,342$ & $-$ & $0.681$ & $0.55$ & $13.9$ \\
      \hline
      MobileNetV2\cite{MobileNetV2}  & $3,568,670$ & $-$ & $0.69$ & $0.59$ & $14.2$ \\
      \hline
      DarkNet-19 \cite{YOLOv2} & $15,770,510$ & $-$ & $0.7$ & $0.61$ & $63.0$ \\
    \end{tabular}
}
     \label{tab:models_acc_det}
    \end{center}

\end{table*} 

\begin{figure}[t]
\centering
\includegraphics[width=0.99\linewidth]{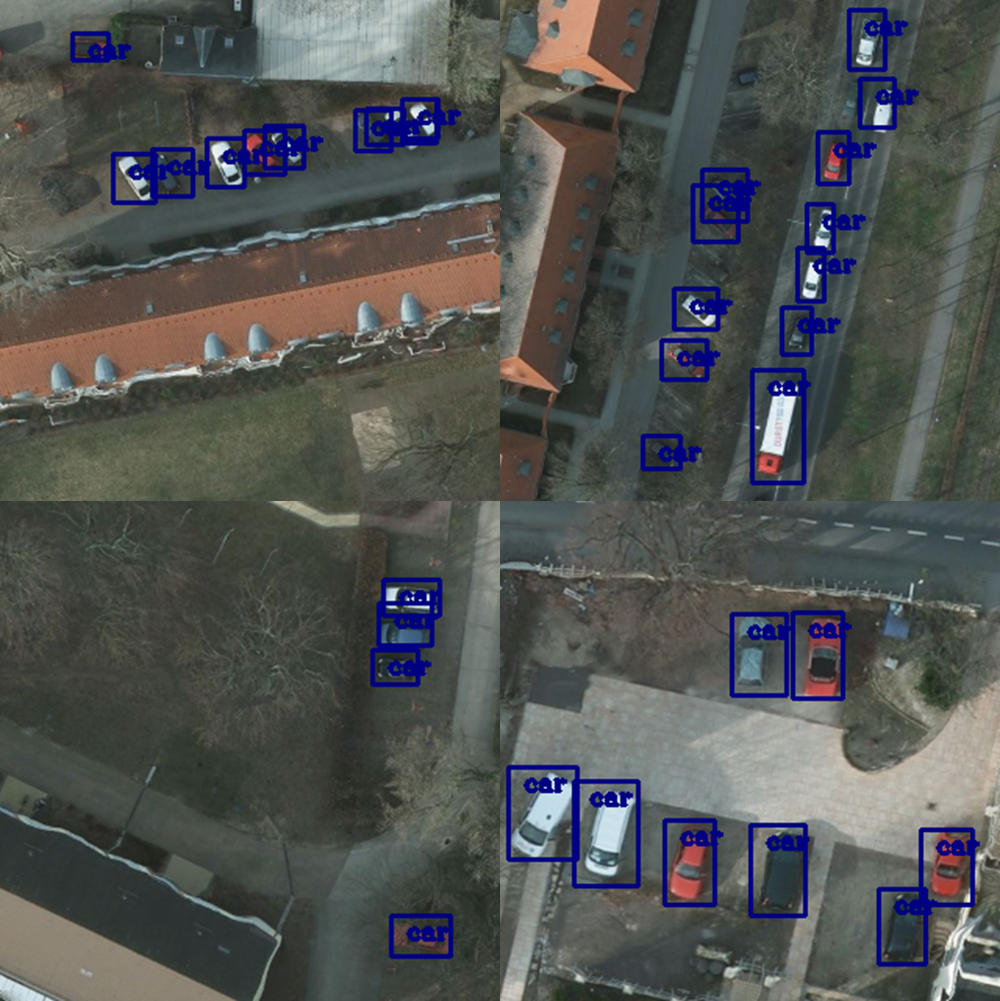}
\caption{Detection results on aerial images for the application of vehicle detection using STeP-Det.}
\label{fig:results}
\end{figure}

\subsection{Object Detection}
\label{sec:ODexp}
The second evaluation scenario illustrates the potential for further enhancements by optimizing the architecture and extending upon the proposed STeP block. In this experiment, the focus is on aerial vehicle detection using Unmanned Aerial Vehicles (UAVs), which serves as a quintessential example of an on-device AI application where the utility of lightweight models is particularly evident \cite{9706868}.

A comparative analysis is conducted involving a network architecture comprised of successive STeP blocks, benchmarked against other lightweight networks featuring ImageNet-pretrained backbones, such as MobileNets \cite{MobileNetV2,howard2017mobilenets} and Darknet-19 \cite{YOLOv2}, as well as a network with LBP convolutional layers from \cite{LBPCOnvNet2017:cvpr}. For the evaluation a dataset comprising approximately $\sim1500$ aerial images is used, with $\sim1000$ images allocated for training and $\sim500$ for testing, all of which are standardized to a size of $320\times320$ pixels. The test dataset consists of approximately $\sim3800$ vehicle instances. The training process extends to $300$ epochs, with an initial learning rate set to $5\times10^{-3}$. Cosine annealing is employed in conjunction with the Adam optimizer \cite{Adam:2014}. Throughout the training phase, both photometric and geometric augmentations are applied to alleviate overfitting. The chosen loss function closely resembles the one utilized in \cite{YOLOv2}. Model performance is evaluated using well-established metrics, including Mean Average Precision (mAP) and Intersection-over-Union (IoU). The architecture of the STeP network adheres to the standard ConvNet structure, as illustrated in Figure \ref{fig:CNN_model}. It features an increasing number of filters with increasing depth and utilizes a combination of striding and average pooling to reduce feature map dimensions. Notably, all convolutional layers integrate the STeP block with ternary patterns. While more sophisticated architectural variations are conceivable, they remain a subject of future research investigation.

An overall comparison is presented in Table \ref{tab:models_acc_det}. As observed from the table, the model employing the proposed backbone consistently achieves a competitive mAP score comparable to networks with significantly higher numbers of parameters. Furthermore, it exhibits superior accuracy compared to the baseline \textit{LBC-Net}, while also attaining a substantially improved Intersection over Union (IoU) score, rivaling that of much larger networks. Notably, both memory usage and parameter count remain significantly lower in comparison to other network architectures. This overall analysis underscores the efficiency of the proposed approach. With a markedly reduced number of trainable parameters ($<< 10^6$) and a conventional network structure, it demonstrates the capability to effectively detect very fine-grained and small dense objects, as shown in Fig. \ref{fig:results}. This accomplishment is particularly noteworthy, given that the task of detecting such objects is widely recognized as challenging, even for more complex network architectures \cite{9706868}.

Overall, the combination of Haar and CS-LBP features yields a diverse set of features that can effectively maintain network performance, while concurrently reducing the computational/storage requirements. Furthermore, the advantages stem without necessitating any modifications to the training procedure. 

\section{Conclusion and Future Work}

This work has demonstrated the effectiveness of utilizing structured ternary patterns, a combination of local binary patterns and Haar features, as a foundation for competitive networks in resource-constrained on-device AI applications. The proposed approach has yielded promising results and has demonstrated performance comparable to that of fully trainable convolutional networks, as validated by the experiments. Distinguishing itself from much of the related work on compression techniques, the proposed method enables direct learning of efficient models. The use of a large number of static parameters in the model offers significant advantages for updating model parameters during deployment and facilitating more efficient federated learning approaches. 

Future research should prioritize the enhancement of both feature and architectural spaces to further optimize designs for maximum efficiency. Additionally, it is imperative to delve deeper into the training process to formulate an optimized training strategy tailored to networks of this nature. Given the reduced number of trainable parameters, consideration should be given to adopting smaller learning rates and potentially extending the number of training epochs. Finally, with the recent trend in using Neural Architecture Search, it would be worthwhile to build a search space based on structured ternary patterns and formulate an automated network design process that also encodes efficiency measures.

\section*{Acknowledgements}
This work is supported by the European Union Horizon 2020 Teaming, KIOS CoE, No. 739551 and from the Government of the Republic of Cyprus through the Deputy Ministry of Research, Innovation, and Digital Policy.\\
The author gratefully acknowledges the support of NVIDIA Corporation with the donation of a A6000 GPU used for this research.

\bibliographystyle{IEEEtran}
\bibliography{references}% common bib file

\end{document}